\documentclass[10pt,twocolumn,letterpaper]{article}

\usepackage{wacv}
\usepackage{times}
\usepackage{epsfig}
\usepackage{graphicx}
\usepackage{amsmath}
\usepackage{amssymb}
\usepackage{multirow}
\usepackage{subcaption}
\usepackage[font=small]{caption}

\wacvfinalcopy 

\def\wacvPaperID{xxxx} 

\ifwacvfinal\fi
\begin{document}

\title{Lightweight 3D Human Pose Estimation Network Training
\\Using Teacher-Student Learning}

\author{Dong-Hyun Hwang
\thanks{This work was done when the first author was on an internship at Clova AI Video, NAVER Corp.}\\
School of Computing\\
Tokyo Institute of Technology\\
{\tt\small hwang.d.ab@m.titech.ac.jp}
\and
Suntae Kim \\
Clova AI Video\\
NAVER Corp.\\
{\tt\small suntae.kim@navercorp.com}
\and
Nicolas Monet\\
Computer Vision Group\\
NAVER LABS Europe\\
{\tt\small nicolas.monet@naverlabs.com}
\and
Hideki Koike\\
School of Computing\\
Tokyo Institute of Technology\\
{\tt\small koike@acm.org}
\and
Soonmin Bae\\
Clova AI Video\\
NAVER Corp.\\
{\tt\small soonmin.bae@navercorp.com}
}

\maketitle
\ifwacvfinal\thispagestyle{empty}\fi
\raggedbottom

\begin{abstract}
We present MoVNect, a lightweight deep neural network to capture 3D human pose using a single RGB camera. 
To improve the overall performance of the model, we apply the teacher-student learning method based knowledge distillation to 3D human pose estimation.
Real-time post-processing makes the CNN output yield temporally stable 3D skeletal information, which can be used in applications directly.
We implement a 3D avatar application running on mobile in real-time to demonstrate that our network achieves both high accuracy and fast inference time.
Extensive evaluations show the advantages of our lightweight model with the proposed training method over previous 3D pose estimation methods on the Human3.6M dataset and mobile devices.
\end{abstract}

\section{Introduction}
We aim to estimate 3D human pose in real-time. Recently human pose estimation has achieved great progress and is being used for sports analytics, body and gesture motion capture in the AR (Augmented Reality) or VR (Virtual Reality) environment.
As VR headset display technology becomes mature, various applications including entertainment, education, and telecommunication are getting released to the market. AR receives even more attention since AR does not require any additional equipment. Nevertheless, creating AR/VR content often requires special settings and devices. We believe that mobile-based marker-less motion capture system will accelerate the advance of AR/VR application market.

Conventional motion capture systems are marker-based relying on wearable suits with sensors and multiple cameras or need depth cameras (e.g. Microsoft Kinect\footnote{https://developer.microsoft.com/en-us/windows/kinect}, Intel RealSense\footnote{https://software.intel.com/en-us/realsense}) to obtain human joint locations. These methods usually require expensive and specialized devices or are restricted to be used in the indoor environment due to calibration procedures and specific light sources required.

\begin{figure}
\begin{center}
\includegraphics[width=\columnwidth]{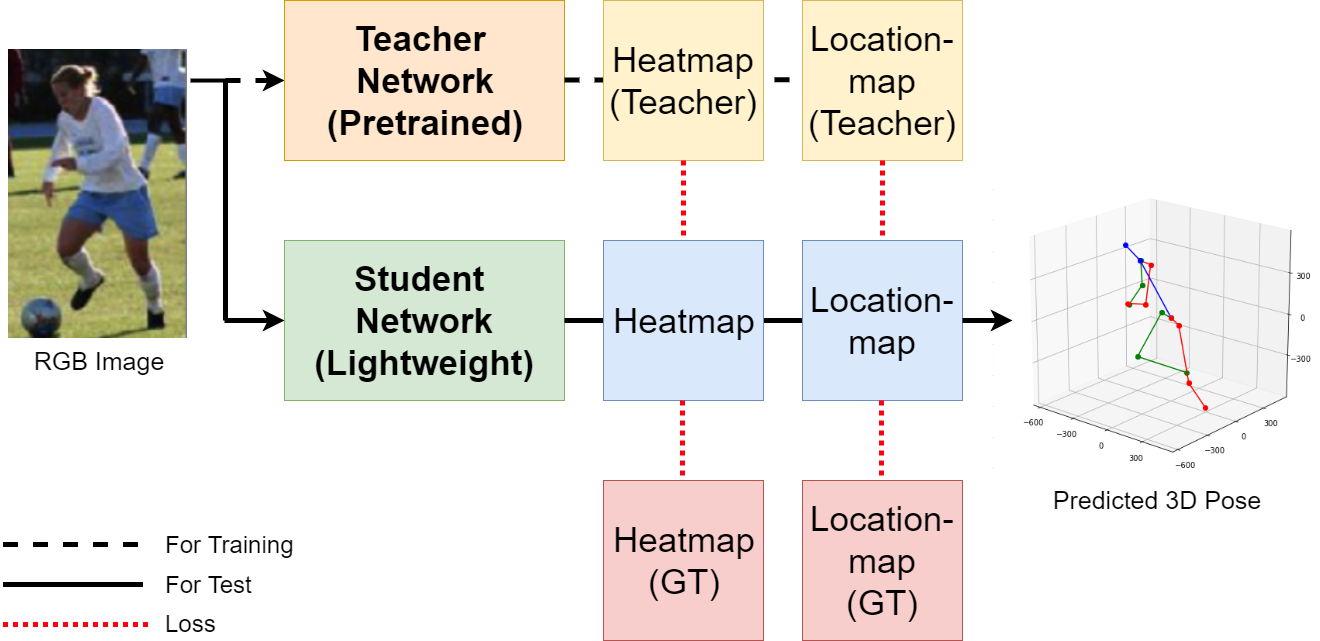}
\end{center}
\caption{An overview of our proposed method. To train lightweight 3D human pose estimation model efficiently, we adopt the basis of knowledge distillation: (1) First, we train a teacher model, which consists of a large number of neural network layers. (2) Then, we train the lightweight model with extra supervision of the teacher model via mimicry loss functions for 3D pose knowledge transfer. The trained lightweight network does not depend on the teacher model and can perform efficient 3D human pose estimation.}
\label{fig:teaser}
\end{figure}
  
Recently, by leveraging the power of deep neural networks, human pose estimation technology with RGB images has been remarkably progressed. However, performance gains of deep learning-based models accompanies high deployment costs due to very deep and wide layers \cite{pmlr-v97-tan19a}. This leads to increased FLOPS (FLoating point OPerations per Second), which is not suitable for devices with limited computing resources such as smartphones or embedded systems. 
To reduce the number of FLOPS, a lightweight model is usually designed with a smaller number of parameters and with efficient operations such as  depthwise convolutions. However, the significantly reduced amount of parameters affect the accuracy of the model. Methods using binarized convolutional neural network (CNN) or quantization \cite{bulat2017binarized, bulat2019improved} often suffer from a lack of generalization capacity.

In this paper, we propose an efficient learning method for 3D human pose estimation model with minimal performance loss while reducing the number of parameters.
We extend the 2D human pose estimation model learning method based on teacher-student learning \cite{Zhang_2019_CVPR} to 3D, and through designing and implementing MoVNect, a lightweight 3D pose estimation model. We observed that the lightweight model trained with the proposed approach achieves higher accuracy than the model trained with the vanilla method.
In addition, we compare the inference time of our model with previous methods running on smartphones and develop an AR application with our model to show the effectiveness of the proposed method.

In summary, our contributions include:
\begin{itemize}
  \item We design MoVNect, a lightweight 3D human pose estimation model that can run in real-time on hardware with limited resources such as smartphones.
  \item We propose a method to efficiently train lightweight 3D human pose estimation with teacher-student learning (Figure \ref{fig:teaser}). The proposed method shows an accuracy improvement of 14\% than the vanilla training method on the Human3.6M test set.
  \item The inference time of various methods on smartphones is evaluated, and the feasibility of the proposed model to be used on various hardware is verified.
  \item We develop a real-time mobile application of 3D avatar with our proposed model to show the practicality of our approach.
\end{itemize}

\section{Related Works}
In this section, we briefly discuss previous approaches for 2D and 3D single human pose estimation using a single RGB camera.

\subsection{2D Human Pose Estimation}
Thanks to advances of deep neural networks, we observe huge improvements on 2D human pose estimation in recent years.
Early approaches extracted features with CNN and directly estimated the joint coordinates as numerical values with fully-connected layers \cite{Jain2013LearningHP,Toshev_2014}.

Later, the heatmap regression-based method \cite{Bulat_2016}, which fully utilizes the spatial context information of the image, was proposed. In this method, a network estimates a heatmap of each joint and indicates the location of joints with the point of the maximum value of each heatmap. Since this method is more accurate than the direct regression method, it has been used in most subsequent approaches \cite{ICCV2017Chen, Chou_2018, Chu_2017, Newell_2016, Sun_2019_CVPR}.

\subsection{3D Human Pose Estimation}
3D human pose estimation is more challenging than 2D estimation because it is an inherently under-constrained problem that requires estimating $z$-axis information not included in two-dimensional images. 
Some early approaches used physics priors \cite{Akhter_2015, Wei_2010} or semiautomatic analysis-by-synthesis fitting of parametric body models \cite{Peng_Guan_2009, Jain_2010}.
3D pose estimation has also been progressed drastically since the utilization of CNN. 
Most studies have attempted to solve the problem in two steps: (1) first estimate 2D joints, and then (2) lift the estimated value into 3D.
There have been attempts to convert 2D joint coordinates into 3D coordinates  \cite{Li_2015, Martinez_2017, videopose3d, Tekin_2017}.
This method is easy to implement, however, the accuracy of 3D output highly depends on the result of the 2D joint prediction. Furthermore, because this method does not utilize the spatial information of CNN layers' outputs, the network has low generalization ability.

Recently, some researches try to combine the two steps. 3D pose estimation methods using volumetric heatmap regression were proposed \cite{Luvizon_2018, Pavlakos_2018, Pavlakos_2017}. However, volumetric heatmaps consumes a lot of memory. To address this issue, a method with stepwise depth resolution increase \cite{Pavlakos_2017} and a soft-argmax based method to find 3D coordinates in low-resolution heatmaps \cite{Luvizon_2018} have been proposed.

The current state-of-the-art methods utilize multi-view geometry to train or inference the network \cite{iskakov2019learnable, epipolar}. These methods achieve accurate pose estimation, however, it requires a lot of computation and memory to process multi-view images.

\begin{figure*}[h!]
\begin{center}
\includegraphics[width=\textwidth]{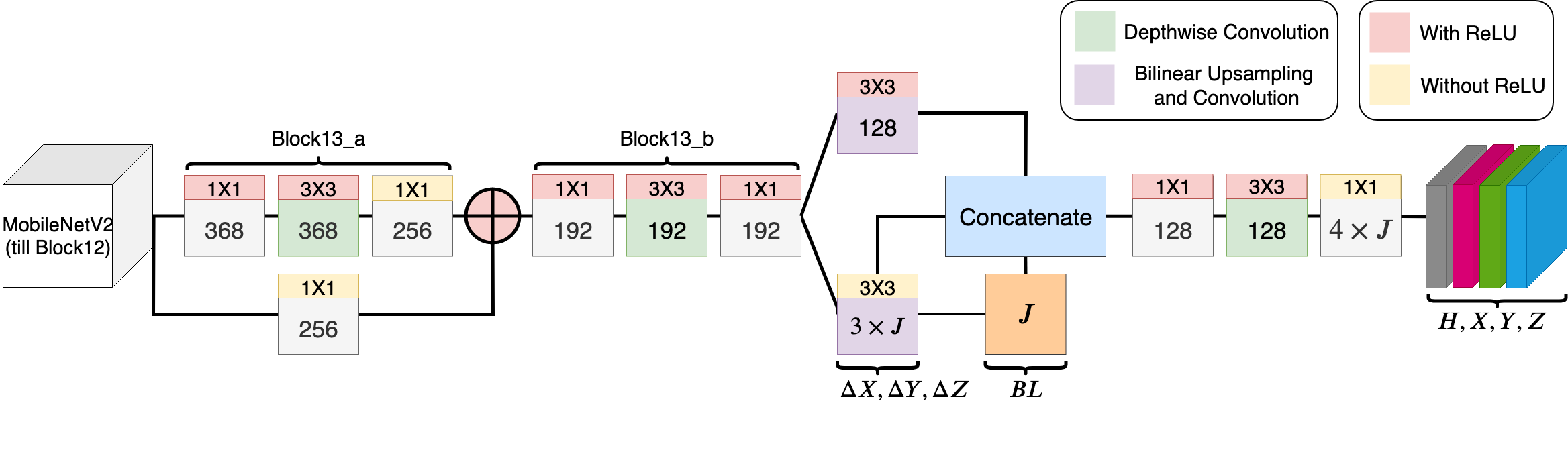}
\end{center}
\caption{Network structure of MoVNect: a single RGB image is fed into the base network (MobileNetV2 till block12), and pointwise and depthwise CNN based structures are used for efficient feature extraction. The intermediate features, $\Delta X$, $\Delta Y$, and $\Delta Z$, are used for bone length-features, auxiliary cue to estimate root-relative 3D human pose. The network predicts heatmaps $H$ and root-relative 3D joint location maps $X, Y, Z$.}
\label{fig:network}
\end{figure*}

Another method regresses location maps where each pixel contains an estimate of a particular coordinate’s value \cite{Mehta_2017}. Because this method uses a two-dimensional location map, it has efficient memory usage and relatively low computational cost, allowing it to run in real-time on a high-end personal computer. Since these features are suitable for our goal, our approach is designed to be based on this method. 

\subsection{Knowledge Distillation}

Knowledge distillation is to transfer the information between different networks with distinct capacities \cite{RichDeep, Bucilua:2006, HintonDist}. The main idea of knowledge distillation is to apply extra supervision using the teacher model in class probabilities \cite{HintonDist}, feature representations \cite{RichDeep,fitnet}, or inter-layer flows \cite{Yim_2017}. It is used for efficient training of small networks difficult to train using large networks, relatively easy to train \cite{fitnet}.
Hinton et al. successfully transferred the knowledge of a large network to a small network \cite{HintonDist}. In addition, methods for online-based distillation \cite{Fukuda_2017, Zhang_2018} are proposed and achieve more effective optimization than previous offline methods.

Recently, there are initial attempts to expand knowledge distillation from classification problems to human pose estimation. The initial attempt estimates human pose using radio signals \cite{Zhao_2018}.
In addition, a method to train an efficient lightweight 2D pose estimation model by knowledge transfer of joint heatmaps \cite{Zhang_2019_CVPR} is proposed and shows significant performance improvement.
Although previous methods show that knowledge distillation can be applied not only to category-level discriminate knowledge but also human pose estimation \cite{Zhang_2019_CVPR, Zhao_2018}, these methods are limited to 2D human pose estimation.
In this work, we propose a knowledge transfer method for 3D human pose networks using teacher-student learning. We also design MoVNect, a lightweight 3D human pose estimation network with the proposed method.
Our lightweight model trained with efficient training method enables accurate pose estimation with very low computation, which can operate on devices with low processing power. 

\section{MoVNect: Lightweight 3D Human Pose Estimation Network}

In 3D human pose estimation, we estimate the 3D pose $P^{3D}$ from a given RGB image $I$. 
$P^{3D} \in R^{3\times J}$ represents the root-relative 3D positions of the $J$ body joints.
We assume our network runs on low power devices (e.g. smartphone, embedded system). Therefore, the network estimates 15 joints ($J=15$), a minimum requirement for the motion of 3D full-body characters.

\subsection{CNN based 3D Pose Regression Network Architecture}
Among previous 3D estimation approaches, the model proposed by Mehta et al. \cite{Mehta_2017} has a good balance between accuracy and inference time. It is easy to apply knowledge distillation because the location map used in the model is 2D spatial information similar to the 2D heatmap. Therefore, we design a lightweight network architecture based on Mehta et al's approach \cite{Mehta_2017} with the model search procedure (session \ref{ch:model_search}) as shown in Figure \ref{fig:network}. 
Our network produces the heatmaps and location maps for all joints $j\in {1..J}$.
We use the till of block 12 of MobileNetV2 \cite{Sandler_2018} as the base network and adopt additional depthwise CNN layers for efficient computation.
We add the bone length-features to the network for an explicit clue to guide the prediction of root-relative location maps as:

$$ BL_{j}=|\Delta X_{j}|+|\Delta Y_{j}|+|\Delta Z_{j}|$$

$\Delta X_{j}$, $\Delta Y_{j}$, and $\Delta Z_{j}$ are intermediate features from our network.
For efficient calculation, bone length-features are calculated using L1 distance instead of L2 distance-based equation proposed by Mehta et al. \cite{Mehta_2017}.
The calculated features are concatenated with other intermediate features and utilized to calculate the final output.

\textbf{Inference:}
we use cropped images based on the person's bounding box when training our network. This makes our network performance affected by the size of the image at runtime. 
To address this issue while maintaining a real time processing on mobile devices, we acquire a bounding box based on the human keypoint $K$, found in the initial few frames of 2D heatmaps with a buffer area 0.2$\times$ the height vertically and 0.4$\times$ the width horizontally. We then track it continuously using previous frames with a momentum of 0.75. To normalize scale, a cropped image based on the bounding box is resized to 256$\times$256 and used as an input to the network.

\subsection{Extra Supervision based on Teacher-Student Learning}
A brief outline of the proposed training method is shown in Figure \ref{fig:teaser}.
Most previous approaches with knowledge distillation are designed for object classification with softmax cross-entropy loss \cite{RichDeep, HintonDist} and not suitable to transfer pose knowledge.
We design mimicry loss functions for 3D pose knowledge transfer based on the method of Zhang et al. \cite{Zhang_2019_CVPR}. 
The network is trained with heatmap loss function $\mathcal{L}_{HM}$ and location map loss function $\mathcal{L}_{LM}$ as

$$
\mathcal{L}_{HM} = {1 \over J}\sum_{j=1}^{J} \{\alpha ||H_{j} - H_{j}^{GT}||_{2} + (1-\alpha )||H_{j} - H_{j}^{T}||_{2}\}
$$
\begin{multline*}
\mathcal{L}_{LM} = \sum_{j=1}^{J} \{\alpha ||H_{j}^{GT} \odot (L_{j}-L_{j}^{GT})||_{2} 
\\ +(1-\alpha )||H_{j}^{GT} \odot (L_{j}-L_{j}^{T})||_{2}\}
\end{multline*}

where $H_{j}$ and $H_{j}^{GT}$ specify the heatmaps for the $j$th joint predicted by the model and ground truth, respectively. 
$\odot$ is the Hadamard product and $L_{j}$ specify the location maps for the $j$th joint predicted by the model . $GT$ and $T$ indicate ground truth and predicted results by the teacher model, respectively.
$\alpha$ is the blending factor between the ground truth and teacher model’s loss terms and set to 0.5.
The teacher-student learning is conducted in each mini-batch and throughout the entire training process. After the training, we only use the student model, already learned with the teacher’s knowledge.

\begin{figure}[h]
\begin{center}
\includegraphics[width=\columnwidth]{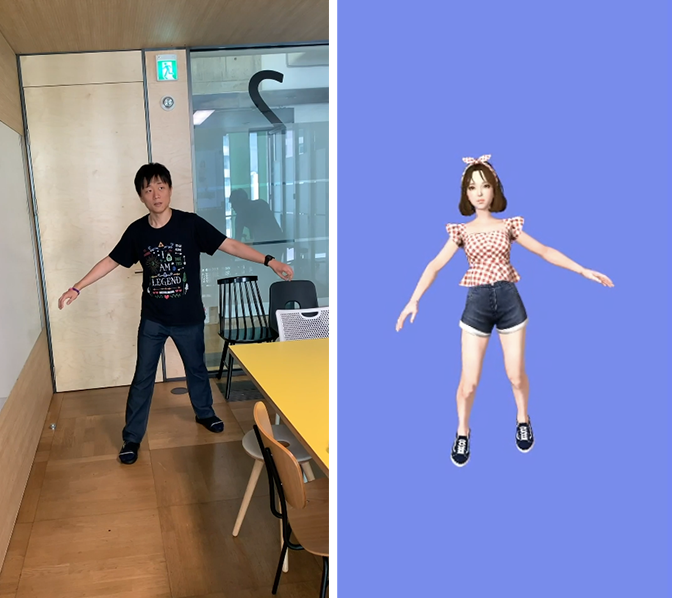}
\end{center}
\caption{3D character control. The processed output can be easily utilized for handling a virtual avatar.}
\label{fig:inter}
\end{figure}

\subsection{Post-processing}
Our model performs CNN based per-frame pose estimation, which leads to a small jitter, an unacceptable artifact in graphics applications.
To reduce this temporal jittering, we apply the 1 Euro filter \cite{oneeuro} to the predicted 2D keypoint and use the filtered keypoint $K$ to refer to the value of the location map. The acquired 3D pose is also filtered to reduce the temporal noise of the prediction results of the continuous images.

The root-relative 3D pose acquired from the cropped image with the bounding box loses the global position information. 
To restore the global position $P_{G}^{3D}$, we use the following simple but effective global pose estimation equation \cite{Mehta_2016}

$$
P_{G}^{3D}=\frac{\sqrt{\sum_{1}^{J}||P_{[xy]}^{j} - \overline{P}_{[xy]}||_{2}}}{\sqrt{\sum_{1}^{J}||K^{j} - \overline{K}||_{2}}} 
\begin{pmatrix}
\overline{K}_{[x]}
\\ \overline{K}_{[y]}
\\ f
\end{pmatrix}
-
\begin{pmatrix}
\overline{P}_{[x]}
\\ \overline{P}_{[y]}
\\ 0
\end{pmatrix}
$$

where $\overline{P}$ and $\overline{K}$ are the 3D, 2D mean over all joints. $P_{[xy]}$ is the $x, y$ part of $P^{3D}$ and single subscripts indicate the particular elements. $f$ is the focal length of the camera.

Since predicted 3D pose is the root-relative 3D position of each joint, it cannot be applied directly for character animations. Hence, inverse kinematics are applied to convert the 3D position of each joint into the orientation and these orientation values are also filtered with the 1 Euro filter \cite{oneeuro}.

In addition, since our model does not have explicit knowledge of the joint angle limits of the human body, our network does not explicitly decline physically invalid poses.
To address this problem, we apply the anatomical joint rotation limits to the calculated angles of each joint to ensure bio-mechanical plausibility.
Through the post processing, our approach exports data directly in a format suitable to 3D character control in real-time as shown in Figure \ref{fig:inter}.

\section{Experiments}
\subsection{Experiment Setup}
We evaluate our model using two measurements:

\textbf{Accuracy:} To measure the accuracy of the model, we use the Human3.6M \cite{h36m_pami} dataset, currently the largest 3D pose dataset. This dataset contains 15 actions performed by 11 subjects. We employ the commonly used evaluation protocol \#1: subject 1, 5, 6, 7, and 8 for training and subject 9 and 11 for testing. Mean Per Joint Position Error (MPJPE) is calculated with the root-relative 3D joint positions from our network.

\textbf{Inference Time:} To confirm the applicability of the proposed lightweight model in the actual mobile environment, we measure inference time on smartphone devices (Apple iPhone series) with a variety of computing hardware specifications (CPU, GPU, NPU). 
We use the Apple Core ML\footnote{https://developer.apple.com/documentation/coreml} framework to convert neural network models to mobile ones and to run these models on the smartphone.

\subsection{Training Details}
Since most 3D human pose datasets consist of indoor images only, the network, trained with only the existing 3D pose dataset, has a lack of generalizability for in-the-wild scenes.
Therefore, following Mehta et al.'s method \cite{Mehta_2016}, we first pre-train the 2D pose estimation using LSP \cite{lsp_dataset} and MPII datasets \cite{mpii_dataset}, and train the 3D pose estimation through Human3.6M \cite{h36m_pami} and MPI-INF-3DHP \cite{Mehta_2016} datasets.
Frames of 3D datasets are sampled with at least one joint movement by $>$200mm between them and cropped using the bounding box of the person.
For the MPI-INF-3DHP dataset, the background augmentation is performed using the Places365 dataset \cite{zhou2017places}, and finally 95k of MPI-INF-3DHP training samples and 100k of Human3.6M training samples are prepared.

We use the Keras \cite{chollet2015keras} framework with the TensorFlow backend for training the network. Some random scaling (0.7-1.0) and gamma correction are performed on training.
RMSProp optimization algorithm \cite{RMSProp} with learning rate to $2.5\times 10^{-4}$ is used for 2D pose training and Adam optimization algorithm \cite{kingma2014adam} with the same learning rate is used for 3D pose training. Mini-batch size is set to 4.
We use the pre-trained base network with ImageNet \cite{imagenet_cvpr09} and batch normalization \cite{Ioffe:2015} before each non-linear activation.

\subsection{Model Search} \label{ch:model_search}
\begin{table}[h]
\centering
\resizebox{\columnwidth}{!}{%
\begin{tabular}{lllc}
\hline
\multicolumn{1}{c}{\multirow{1}{*}{Network}} & \multicolumn{2}{c}{Network Structure}                           & \multirow{1}{*}{Upsampling Method} \\ \cline{2-3}
\multicolumn{1}{c}{}                         & \multicolumn{1}{c}{Block13\_a} & \multicolumn{1}{c}{Block13\_b} &                                    \\ \hline
Type A                                       & 368, 368, 256                  & 192, 192, 128                  & Bilinear + Conv2D                  \\
Type B                                       & 368, 368, 256                  & 192, 192, 128                  & TransposedConv2D                   \\
Type C                                       & 512, 512, 512                  & 256, 256, 128                  & Bilinear + Conv2D                  \\
\hline
\end{tabular}%
}
\caption{Specification for our prototype MoVNect models. Sequential numbers on Network Structure column denote the number of CNN layers, which make up each block.} \label{tab:model_1}
\end{table}

To find a suitable model, which has a good balance between accuracy and inference time, we design and train various types (Type A, B, C) of models that have a different number of layers on Block13\_a, Block13\_b, and upsampling method (Bilinear upsampling + Convolution, Transposed Convolution). 
See Table \ref{tab:model_1} for specification of our prototype MoVNect networks.

\begin{table}[h]
\centering
\resizebox{\columnwidth}{!}{%
\begin{tabular}{lllcc}
\hline
\multicolumn{1}{c}{\multirow{1}{*}{Network}} & \multicolumn{2}{c}{Network Structure}  & \multirow{1}{*}{\# Param} & \multirow{1}{*}{MPJPE} \\ \cline{2-3}
\multicolumn{1}{c}{}                         & \multicolumn{1}{c}{Block13\_a} & \multicolumn{1}{c}{Block13\_b} &                                                                                  &                        \\ \hline
Type A                                       & 368, 368, 256                  & 192, 192, 128                  & \textbf{1.03M}                                                                            & 113.3                  \\
Type C                                       & 512, 512, 512                  & 256, 256, 128                  & 2.69M                                                                            & \textbf{108.2}                  \\ \hline
\end{tabular}%
}
\caption{Performance analysis with the number of layers. Metric: average MPJPE(mm). M:$10^{6}$.} \label{tab:model_2}
\end{table}

First, we measure the performance and inference time correlation with the number of layers. we design MoVNect-Small (Type A), MoVNect-Large (Type C) and measure the average MPJPE on the test set of Human3.6M as shown in Table \ref{tab:model_2}. Because of the deep neural network's suboptimal trade-off between the representation capability and the computational cost,
Type C has no significant improvement in accuracy (about 5mm improvement), even though the number of parameters in the network is twice that of Type A.

\begin{table}[h]
\centering
\begin{tabular}{cccc}
\hline
Network & Upsampling Method & \multirow{1}{*}{\# Param} & MPJPE \\ \hline
Type A  & Bilinear + Conv2D & \textbf{1.03M} & \textbf{113.3}                   \\
Type B  & TransposedConv2D  & 1.13M & 126.7                      \\ \hline
\end{tabular}%
\caption{Performance analysis with upsampling methods. Metric: average MPJPE(mm). M:$10^{6}$} \label{tab:model_3}
\end{table}

Next, we measure the change in accuracy according to the upsampling method which increases the resolution of the network output. As shown in Table \ref{tab:model_3}, we compare Type A and Type B, which use bilinear upsampling with convolution and transposed convolution, respectively.
According to the results, although transposed convolution method (Type B) requires more parameters, accuracy was lower than resize-convolution method  (Type A). 
We presume that while transposed convolution method has a unique entry for each output window, resize-convolution method is implicitly weighted in a way that it reduces the high frequency artifacts.
Based on these results, we finally choose Type A network for MoVNect.

\begin{table*}[ht!]
\begin{minipage}{\textwidth}
\setlength{\tabcolsep}{1pt} 
\resizebox{\textwidth}{!}{%
\begin{tabular}{llllllllllllllll|c|c}
\hline
Methods                                                        & Direct. & Discuss & Eating & Greet & Phone & Photo & Pose  & Purch. & Sitting & SittingD. & Smoke  & Wait  & WalkD. & Walk & WalkT. & Avg. & \# Param \\ \hline
Zhou et al.\cite{zhou2016}                                                            & 87.4    & 109.3   & 87.1   & 103.2 & 116.2 & 143.3 & 106.9 & 99.8   & 124.5   & 199.2     & 107.4  & 118.1 & 114.2  & 79.4 & 97.7   & 113.0  & - \\
Du et al.\cite{Du_2016}                                                            & 85.1    & 112.7   & 104.9  & 122.1 & 139.1 & 135.9 & 105.9 & 166.2  & 117.5   & 226.9     & 1120.0 & 117.7 & 137.4  & 99.3 & 106.5  & 126.5 & - \\
Park et al.\cite{Park_2016}.                                                            & 100.3   & 116.2   & 90.0   & 116.5 & 115.3 & 149.5 & 117.6 & 106.9  & 137.2   & 190.8     & 105.8  & 125.1 & 131.9  & 62.6 & 96.2   & 117.3 & - \\
Mehta et al.\cite{Mehta_2016}                                                           & 52.6    & 63.8    & 55.4   & 62.3  & 71.8  & 52.6  & 72.2  & 86.2   & 120.6   & 66.0      & 79.8   & 64.0  & 48.9   & 76.8 & 53.7   & 68.6 & - \\
Martinez et al.\cite{Martinez_2017} w/ SH                                                        & 51.8    & 56.2    & 58.1  & 59.0  & 69.5  & 78.4  & 55.2  & 58.1   & 74.0    & 94.6 & 62.3   & 59.1  & 65.1   & 49.5 & 52.4   & 62.9 & 19.3M \\
Mehta et al.\cite{Mehta_2017}                                                           & 62.6    & 78.1    & 63.4   & 72.5  & 88.3  & 63.1  & 74.8  & 106.6  & 138.7   & 78.8      & 93.8   & 73.9  & 55.8   & 82.0 & 59.6   & 80.5 & 14.6M \\
Pavlakos et al.\cite{Pavlakos_2018}                                                        & 48.5    & 54.4    & 54.4   & 52.0  & 59.4  & 65.3  & 49.9  & 52.9   & 65.8    & 71.1      & 56.6   & 52.9  & 60.9   & 44.7 & 47.8   & 56.2  & - \\
Yang et al.\cite{Yang_2018}                                                           & 51.5    & 58.9    & 50.4   & 57.0  & 62.1  & 65.4  & 49.8  & 52.7   & 69.2    & 85.2      & 57.4   & 58.4  & 43.6   & 60.1 & 47.7   & 58.6  & - \\
Kocabas et al.\cite{epipolar}           & -    & -    & -   & -  & -  & -  & - & - & - & - & - & -  & -  & - & -   & \textbf{51.8}  & 34M \\ \hline
\textbf{Ours}  & 80.6    & 96.3   & 92.2  & 90.4   & 116.1  & 82.1 & 110.9   &188.4  & 224.6  & 106.9    & 123.2   & 98.9  & 90.4   & 117.3  & 80.5   & 113.3 & \textbf{1.03M}      \\
\textbf{Ours$\dagger$} & 72.4    & 83.4   & 76.9  & 82.1   & 101.9  & 70.4 & 91.8   &156.5  & 193.0  & 92.8    & 108.4   & 85.1  & 76.8   & 97.2  & 70.5   & 97.3 \textbf{(14\%$\downarrow$)} & \textbf{1.03M}      \\
\hline
\end{tabular}
}
\vspace{-1em} 
\caption{Results of our network's raw CNN predictions. All frames of subject 9 and 11, cropped with the ground truth bounding box, were used for evaluation. $\dagger$ means the model trained with the proposed teacher-student learning method. Metric: MPJPE(mm). M:$10^{6}$.} \label{tab:h36m}
\end{minipage}
\vspace{1em} 

\begin{minipage}{\textwidth}
\centering
\resizebox{\textwidth}{!}{%
\begin{tabular}{l|l|l|l|ll|ll|lll|lll}
\hline
\multicolumn{1}{c|}{Methods} & \multicolumn{3}{c|}{Cost-Effectiveness} & \multicolumn{10}{c}{Inference Time on Devices} \\ \cline{2-14} 
\multicolumn{1}{c|}{} & \multicolumn{1}{c|}{MPJPE} & \multicolumn{1}{c|}{\# Param} & \multicolumn{1}{c|}{FLOPS} & \multicolumn{2}{c|}{iPhone7} & \multicolumn{2}{c|}{iPhone 8} & \multicolumn{3}{c|}{iPhone X} & \multicolumn{3}{c}{iPhone XS} \\
\multicolumn{1}{c|}{} & \multicolumn{1}{c|}{} & \multicolumn{1}{c|}{} & \multicolumn{1}{c|}{} & \multicolumn{1}{c}{CPU} & \multicolumn{1}{c|}{GPU} & \multicolumn{1}{c}{CPU} & \multicolumn{1}{c|}{GPU} & \multicolumn{1}{c}{CPU} & \multicolumn{1}{c}{GPU} & \multicolumn{1}{c|}{NPU} & \multicolumn{1}{c}{CPU} & \multicolumn{1}{c}{GPU} & \multicolumn{1}{c}{NPU} \\ \hline
Mehta et al.\cite{Mehta_2017} & 80.5 & 14.6M & 7.3M & 275 & 175 & 215 & 140 & 270 & 120 & 120 & 200 & 110 & 17 \\
Martinez et al.\cite{Martinez_2017} w/ SH & 62.9 & 19.3M & 22M & 750 & 200 & 300 & 160 & 350 & 160 & 160 & 270 & 120 & 20 \\
Kocabas et al.\cite{epipolar} & \textbf{51.8} & 34M & 14M & 500 & 220 & 210 & 210 & 230 & 160 & 160 & 200 & 125 & 50 \\ \hline
\textbf{Ours} & 97.3 & \textbf{1.03M} & \textbf{1.35M} & \textbf{48} & \textbf{56} & \textbf{40} & \textbf{33} & \textbf{37} & \textbf{28} & \textbf{28} & \textbf{32} & \textbf{22} & \textbf{6} \\ \hline
\end{tabular}%
}
\vspace{-1em} 
\caption{Comparison of networks' cost-effectiveness and inference time on mobile devices with various hardware configurations. Metrics: average MPJPE(mm), the number of parameters, FLOPS, and average inference time(ms). M:$10^{6}$.} \label{tab:throughput}
\end{minipage}
\end{table*}

\section{Results}
\subsection{Accuracy Results on Human3.6M Dataset}

Our results on Human3.6M are shown in Table \ref{tab:h36m}. Our model shows competitive accuracy compared with other methods. In particular, the model trained with teacher-student learning (marked with $\dagger$) shows significantly improved accuracy (14\% average MPJPE reduction). Even though our model consists of a very small number of parameters, it has cost-effective accuracy.
These results show that our proposed training approach has good generalization capability in yielding cost-efficient 3D pose estimation models. 

We compare the computation amounts of networks in Table \ref{tab:throughput} (see column 2).
Compared with the teacher model \cite{Mehta_2017}, our model only requires 7.1\% (1.03M / 14.6M) parameters and 18.5\% (1.35M / 7.3M) computational amount but achieves 82.7\% (97.3 / 80.5) accuracy in average MPJPE.
When compared with the best performer \cite{epipolar}, our model with 3\% (1.03M / 34M) parameters and 9.6\% (1.35M / 14M) computational amount achieves 53.2\% (97.3 / 51.8) accuracy.
Our model with the proposed method has cost-effectiveness advantages over other alternative models. 
Note that we apply the teacher-student learning method without changing any network structure. 
Based on the results in Table \ref{tab:model_2}, we presume that several times more parameters with additional layers are required to overcome the performance gap without our teach-student learning.
This design choice is quite critical and inevitable in real-time applications.

\begin{figure*}[th]
\begin{center}
\includegraphics[width=\textwidth]{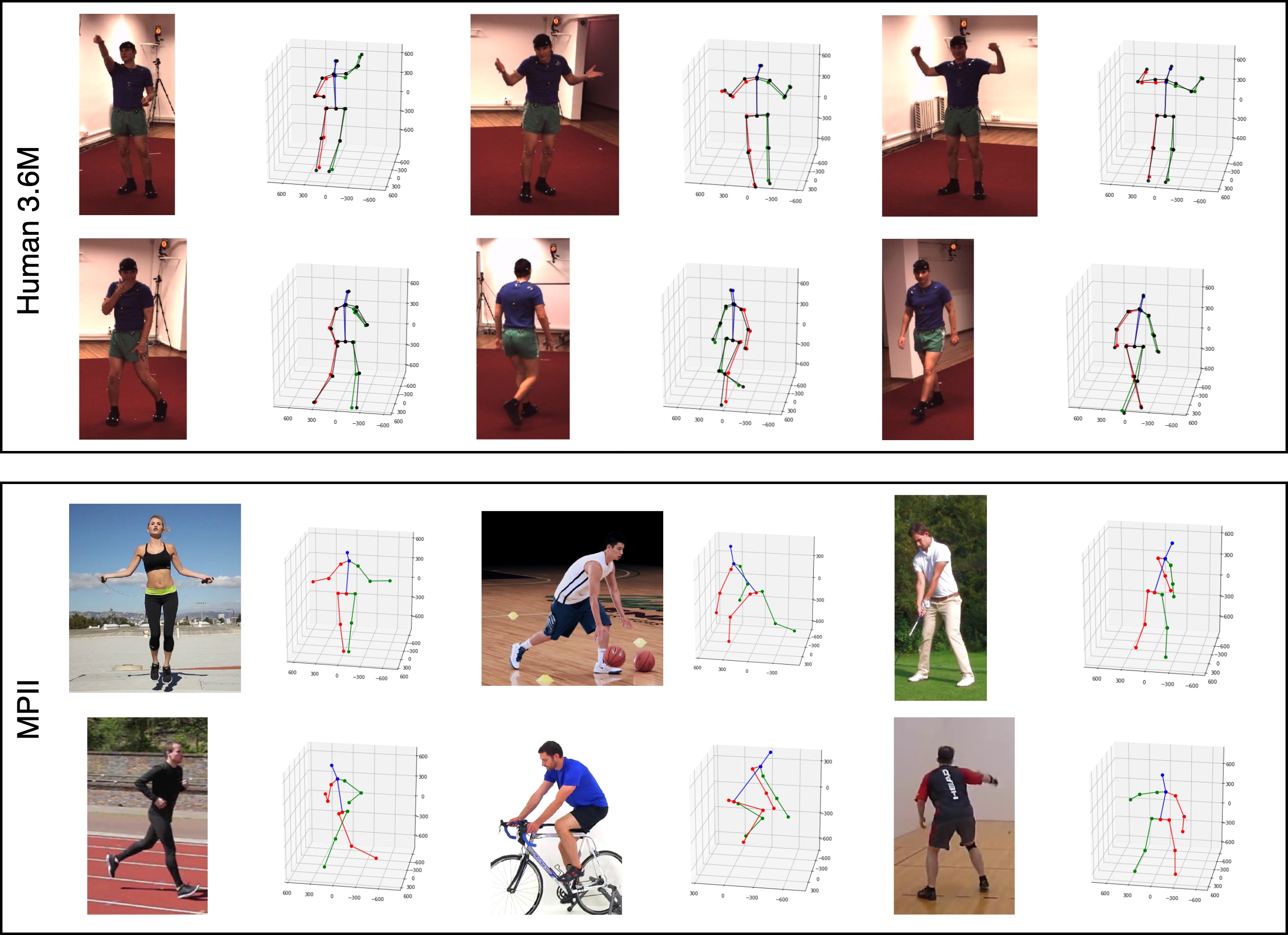}
\end{center}
\caption{Qualitative results on the test set of Human3.6M(3D) and MPII(2D) datasets.
Left: the input images; Right: the results of 3D pose prediction from a different viewpoint, the black skeleton is the ground truth of the Human3.6M dataset.}
\label{fig:qualitative}
\end{figure*}

In Figure \ref{fig:qualitative}, we show qualitative results on Human3.6M and MPII datasets to demonstrate the generalization of our network to general scenes.

\subsection{Inference Time Benchmark Results on Mobile Devices} \label{sec:throughput}

Table \ref{tab:throughput} (see column 3) shows the inference time benchmark results on mobile devices.
Throughout all the devices we test, our model’s inference time outperforms other networks.
Even with low-end devices (iPhone 7 with CPU), our model runs out over 20fps and with high-end devices (iPhone XS with NPU), the throughput reaches over 160fps. Note that except our model, which has low FLOPS and memory consumption, there is no other method that can perform over 10fps on CPU and GPU.
Compared to Mehta et al.’s model \cite{Mehta_2017}, which is used as the teacher model, our model performs at least 283\% (iPhone XS with NPU) and up to 730\% (iPhone X with CPU) faster throughput.
Compared to the best performer \cite{epipolar}, our model shows at least 393\% (iPhone 7 with GPU) and up to 1042\% (iPhone 7 with CPU) faster inference time.

As the device processing power increases, the difference in throughput between networks decreases. In particular, in the case of utilizing a dedicated neural network accelerator such as NPU, all models of the comparison group are able to process more than 20 fps.
However, different from other networks, our network does not have a huge gap across different processing unit types.
Hence, if the model inference is done by CPU in low-end devices and by NPU in high-end devices, the GPU could be fully utilized for graphic rendering, and this is a great advantage for CG applications.
Furthermore, most users do not have smartphones equipped with a dedicated processor for neural networks. Our proposed approach is expected to contribute to the spread of deep learning-based interactive applications until high-end devices are deployed broadly.

\subsection{Applications}
Our proposed network can be applied for various interactive mobile applications because it can provide motion data in a format suitable to 3D avatar control in real-time on mobile devices. In addition, since our network has low inference time, enough times remains for CG rendering for the application. 

\begin{figure}
\begin{center}
\includegraphics[width=\columnwidth]{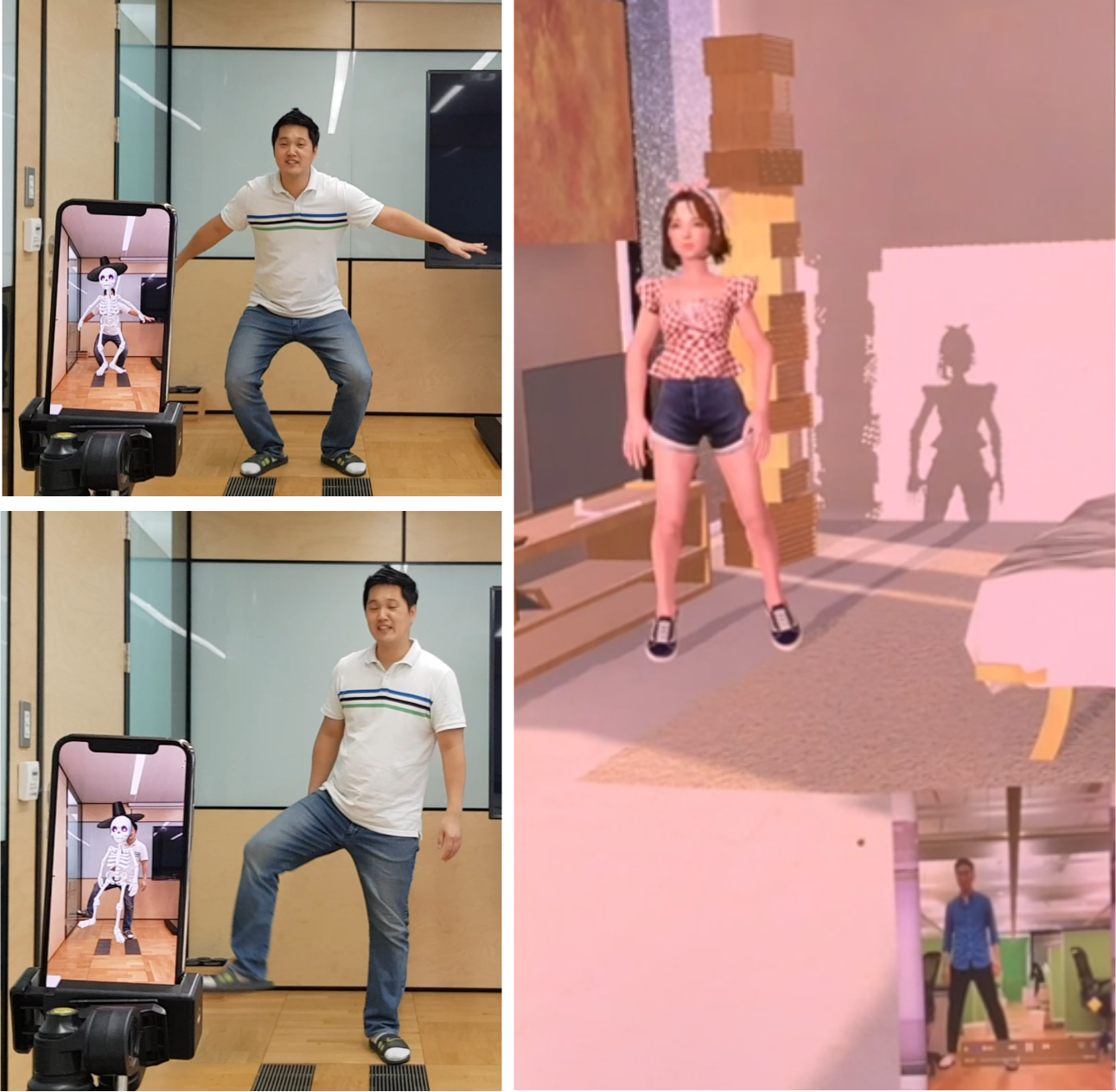}
\end{center}
\caption{AR-based real time 3D avatar mobile application. Our lightweight network can be utilized for interactive applications, which provide immersive experiences to users.} 
\label{fig:app}
\end{figure}

\textbf{Augmented and Virtual Reality:} Smartphone, which has built-in camera, inertial measurement unit sensor, and display, is the best portable device for AR and VR applications. Our method enables applications that provide the user with immersive content through a virtual avatar of the user exactly mimicking the user's real pose using a single RGB camera as shown in Figure \ref{fig:app}.
It also enables a real time interaction in body gesture capturing applications.

\textbf{Motion Capture Simply Accessible:} 
Our lightweight network can be used in a variety of devices that have low-computation power. Without communication with a high-performance server for processing algorithms, the network can be deployed and run directly on various mobile IoT devices in our daily life and can be applied in various real-life scenarios such as interactions with objects through body gestures, healthcare, and so on. 
For example, our algorithm can be used to recognize body language or to analyze walking postures of the elderly with common smart home devices.

\section{Discussion and Future Work}
To the best of our knowledge, our training approach is the first knowledge transferring method for 3D human pose estimation networks. 
MoVNect achieves a well-balanced performance between accuracy and inference time. 
Nevertheless, it still has certain limitations that can be addressed in future work.
In this paper, we transfer the knowledge to the lightweight network based on the location map, thanks to the similar output type to the 2D pose network. 
Furthermore, because the mimicry loss function is very simple, we envision that we can easily apply our knowledge transfer method to various 3D human pose networks.

We have focused on estimating the 3D pose of a single person, which can run in real-time across various devices. Currently, latest high-end smartphones tend to be equipped with dedicated accelerators such as NPU. The proposed fully-convolutional network could be scaled to multiple persons if such devices have enough computational capacity.

To reduce the resource and power consumption, most mobile deep learning frameworks do not fully support recurrent architectures. We also design our network based on per-frame prediction and this may lead to some temporal instability, similar to previous per-frame prediction approaches. 
We believe that our post-processing method should reduce temporal jitters enough to be usable and practical in various fields. 
Furthermore, in the near future, mobile devices will have more processing power and we will be able to expand per-frame to video processing levels using a recurrent approach.

In addition, to reduce inference time, our network uses a single scale of the cropped image. Processing each frame inference with multiple scales of the image (scale-space search) makes it difficult to guarantee real-time performance on low power devices. For applications that require a better accuracy for the pose, two different scales (like 0.7 and 1.0) of the cropped image can be used.

\begin{figure}
\begin{center}
\includegraphics[width=\columnwidth]{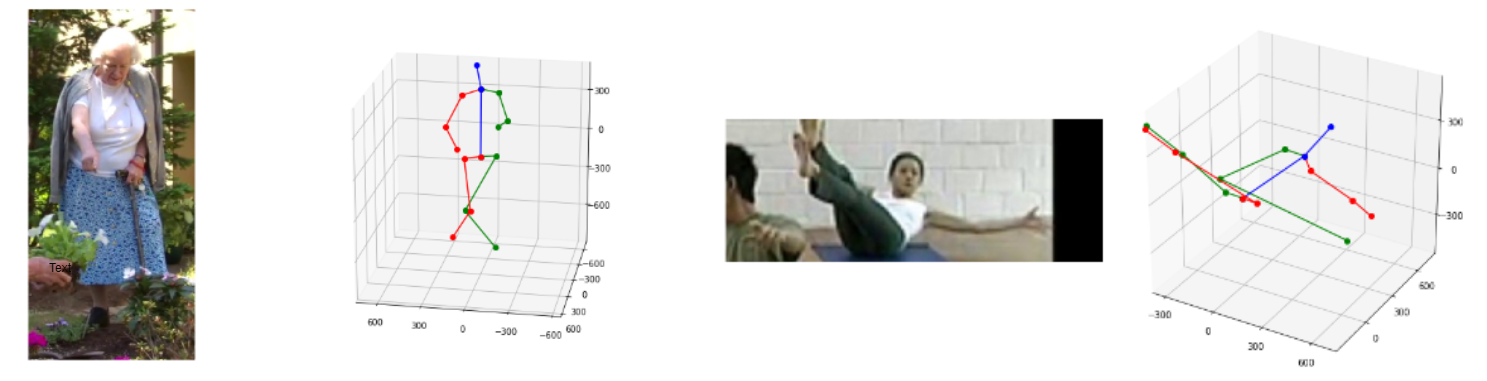}
\end{center}
\caption{Failure cases of our model. Left: knees are crossed because of body part occlusion. Right: the position of the right hand is mislocated to the left hand because the right hand is occluded with the extreme pose.}
\label{fig:fcase}
\end{figure}

A few failure cases are illustrated in Figure \ref{fig:fcase}. 
Our location map-based approach relies on 2D heatmap detection results and our lightweight model is not robust enough to occlusion.
As future work, we will apply a pose encoding-decoding scheme \cite{mehta2019xnect}, robust to occlusion, to our network.
Despite these limitations, we observe that our method proposes an initial step in the direction of a training method for efficient lightweight 3D motion capture.

\section{Conclusion}
In this paper, we propose MoVNect, a lightweight 3D human pose estimation model, and an efficient training strategy based on teacher-student learning.
We make the step from existing 2D pose estimation with knowledge distillation to 3D pose estimation.
Moreover, we present extensive evaluations on human pose and inference time benchmarks. Based on the results, we observe that our proposed teacher-student learning method significantly improves the accuracy of the model, and our network trained with the proposed method achieves very fast inference time with reasonable accuracy on various devices from low-end to high-end.
We demonstrate these advantages on real mobile devices with an AR-based 3D avatar application.
We hope that this work would act as an ignition of efficient training methods for lightweight neural networks in 3D human pose estimation.

{\small
\bibliographystyle{ieee}
\bibliography{main}
}

\end{document}